%% file: conference_101719.tex
\documentclass[conference]{IEEEtran}
\IEEEoverridecommandlockouts
\usepackage{cite}
\usepackage{amsmath,amssymb,amsfonts}
\usepackage{algorithmic}
\usepackage{graphicx}
\usepackage{textcomp}
\usepackage{xcolor}
\usepackage{float}
\def\BibTeX{{\rm B\kern-.05em{\sc i\kern-.025em b}\kern-.08em
    T\kern-.1667em\lower.7ex\hbox{E}\kern-.125emX}}

\begin{document}

\title{Interpretable Dual-Stream Learning for Local Wind Hazard Prediction in Vulnerable Communities
}

\author{
    \IEEEauthorblockN{
    Mahmuda Akhter Nishu\IEEEauthorrefmark{1}, 
    Chenyu Huang\IEEEauthorrefmark{2},
    Milad Roohi\IEEEauthorrefmark{3},
    Xin Zhong\IEEEauthorrefmark{1}
    }
    
    \smallskip
    \IEEEauthorblockA{
    \IEEEauthorrefmark{1}
    Department of Computer Science, University of Nebraska Omaha, Omaha, NE, USA
    \\
    \IEEEauthorrefmark{2}
    Aviation Institute, University of Nebraska Omaha, Omaha, NE, USA
    \\
    \IEEEauthorrefmark{3}
    Durham School of Architectural Engineering and Construction, University of Nebraska Lincoln, Omaha, NE, USA
    \\ 
    mnishu@unomaha.edu, chenyuhuang@unomaha.edu, milad.roohi@unl.edu, xzhong@unomaha.edu
    }
}

% \and
% \and
% \and
% \and
% \and

\maketitle

\begin{abstract}
\input{sections/abstract}

\end{abstract}


\section{Introduction}
\label{sec: intro}
\input{sections/introduction}

\section{Related Work}
\label{sec: literature}
\input{sections/related}

\section{Proposed Method}
\label{sec: proposed method}
\input{sections/method}

\section{Experimental Results}
\label{sec: experiments}
\input{sections/experiments}

% \vspace{-1.0em}
\section{Conclusion}
\label{sec: conclusion}
\input{sections/conclusion}

\bibliographystyle{IEEEtran}
\bibliography{ref}

\end{document}

%% file: sections/abstract.tex
% Problem + Impact Area
Wind hazards such as tornadoes and straight-line winds frequently affect vulnerable communities in the Great Plains of the United States, where limited infrastructure and sparse data coverage hinder effective emergency response. 
% Gap Summary
Existing forecasting systems focus primarily on meteorological elements and often fail to capture community-specific vulnerabilities, limiting their utility for localized risk assessment and resilience planning. 
% Proposed Work
To address this gap, we propose an interpretable dual-stream learning framework that integrates structured numerical weather data with unstructured textual event narratives. Our architecture combines a Random Forest and RoBERTa-based transformer through a late fusion mechanism, enabling robust and context-aware wind hazard prediction. 
% Contributions
The system is tailored for underserved tribal communities and supports block-level risk assessment. Experimental results show significant performance gains over traditional baselines. Furthermore, gradient-based sensitivity and ablation studies provide insight into the model’s decision-making process, enhancing transparency and operational trust. 
% Broader Value
The findings demonstrate both predictive effectiveness and practical value in supporting emergency preparedness and advancing community resilience.

%% file: sections/introduction.tex
 Wind hazards in the Great Plains of the United States, such as tornadoes and straight-line winds, remain among the most destructive and fast-moving natural disasters, often leading to widespread infrastructure damage and threats to public safety. Current weather forecast services mainly provide predictions for individual meteorological elements. Although meteorological forecasting systems have improved, they frequently lack the localized precision without considering community characteristics and vulnerabilities required to support timely and community-specific responses. These limitations are particularly prominent for underserved regions—such as tribal communities—where infrastructure is often fragile, local weather observational data and historical records are usually sparse, and the natural disaster response and recovery resources are limited~\cite{howard2019tribal, mcallister2013developing, ortiz2002tribal, quick202116}. In addition, effective risk prediction requires not only wind hazard forecasts but also up-to-date intelligence on infrastructure inventory, population, and asset vulnerabilities. As a result, underserved communities are usually more vulnerable to wind hazards, resulting in a gap in community-specific wind hazard risk prediction.

% what we propose? tech
To address these interdisciplinary challenges in wind hazard risk assessment, this study proposes an interpretable, AI-driven framework for localized wind hazard prediction. The system integrates structured numerical weather data with descriptive textual event narratives to assess the likelihood and severity of wind hazards at the community level. By combining physical measurements with human-observed conditions, the model captures both atmospheric trends and situational context, enabling more informative and actionable predictions than traditional single-source approaches. 
% what we propose? app
The framework is tailored for application in tribal communities across the United States Midwest, with the goal of supporting block-level or household-level hazard assessments. These outputs are designed to inform emergency preparedness, real-time warning decisions, and the prioritization of response resources in underserved regions. By emphasizing local context, operational adaptability, and community relevance, the framework lays the groundwork for integration into decision-support platforms such as GIS dashboards and emergency planning tools.

Our contributions are threefold. 
(1) We develop a deployable, AI-driven framework to support localized wind hazard risk prediction for underserved tribal communities, bridging the gap between traditional weather forecasts focusing on meteorological conditions and the practical needs of community-specific emergency management that must consider community characteristics and vulnerabilities against wind hazards. 
(2) We introduce a novel dual-stream deep learning architecture that integrates structured numerical weather data with unstructured descriptive event narratives through a fusion-based design, enabling the model to capture both atmospheric patterns and contextual signals associated with wind hazards. 
(3) We enhance interpretability through gradient-based and ablation-driven feature importance analysis, facilitating transparent decision support and practical integration into emergency management workflows.

%% file: sections/related.tex
\subsection{Community Resilience Development of Underserved Communities}

The development of resilience within underserved communities has presented challenges due to limitations in technology access, the availability of a skillful workforce, and the mechanisms for effective technology integration and implementation~\cite{howard2019tribal}. Recent devasting events underscore the urgent need to boost resilience through enhanced preparedness and infrastructure planning. In recent years, there have been significant advancements in computational wind resilience modeling of communities, including NSF NHERI-funded SimCenter tools for individual systems and NIST-funded Interdependent Networked Community Resilience Modeling Environment (IN-CORE) platform for community-level modeling~\cite{mcallister2013developing}. However, there are various challenges associated with this process that motivate this study, including i) data availability and community-specific characteristics, ii) the accuracy and accessibility of wind hazard prediction models for United States Midwest underserved communities, and iii) the integration of wind hazard prediction with unique community vulnerabilities which may substantially differ by regions~\cite{ortiz2002tribal, quick202116}. 
Analysis of the current landscapes of research and practice in disaster management, community resilience, and relevant fields reveals the challenges in protecting communities from extreme wind hazards in the Great Plains. To facilitated effective community resilience improvement and the development of best practice of emergency management against wind hazard, a good understanding of how to translate the knowledge of meteorological conditions and community vulnerabilities into effective wind hazard risk prediction is expected to be instrumental.
 
The concept of community resilience is particularly significant for vulnerable populations, such as those residing on tribal lands, where inadequate infrastructure and limited resources exacerbate the risks posed by extreme weather events. For these communities, improving resilience against wind hazards is essential due to unique socio-political and logistical challenges that demand precise, real-time, and localized data to effectively mitigate risk. One of the primary challenges that tribal communities face during severe weather is the lack of localized hazard predictions based on actual community characteristics and difficulties in securing adequate resources for disaster mitigation, preparation, response, and recovery. 
Historically, tribal governments have struggled to access necessary disaster funding due to a lack of technological infrastructure and complex documentation requirements for recording losses, which further exacerbated the infrastructure conditions in those communities and made them more vulnerable for less severe wind events~\cite{tsai2024co, huang2023research}. It is urgent to develop practical tools to predict wind hazard risk considering the actual community characteristics for so many underserved communities. 
Gupta et al. explored the potential of AI-based tools to enhance emergency management systems for tribal lands and highlighted the benefits of real-time data collection through AI-powered platforms, which could improve resource allocation, communication, and decision-making during emergencies~\cite{gupta2024utilizing}.

\subsection{AI-Driven Models for Wind Hazard Prediction}

Wind hazards, including tornadoes and straight-line winds, pose serious threats to both urban and rural communities, particularly in underserved regions where infrastructure and early warning systems are limited. Traditional wind hazard prediction models rely heavily on numerical indicators such as wind speed, air pressure, and precipitation. While effective for broad-scale forecasting, these approaches often lack the granularity needed for localized risk assessment and rapid emergency response. Recent studies have highlighted the limitations of conventional methods and emphasized the importance of incorporating heterogeneous data sources to enhance both precision and contextual relevance \cite{chang2014literature, ma2022hybrid, huang2023research}.

Advancements in machine learning have enabled the development of more sophisticated hazard models that integrate structured numerical data with unstructured sources such as textual weather narratives. Lin et al. surveyed machine learning techniques for wind forecasting and demonstrated the potential of ensemble models and neural networks to improve performance~\cite{lin2021wind, pianforini2024real}. These findings are echoed in recent work by Qiu et al. and Bentsen et al., who explored hybrid and spatiotemporal deep learning models capable of leveraging multiple data modalities for greater predictive accuracy and spatial resolution \cite{qiu2024windformer, bentsen2023spatio}. 
In parallel, transformer-based architectures—originally developed for natural language processing—have shown increasing promise in meteorological applications. For example, the BERT model~\cite{kenton2019bert} and its derivatives have been leveraged in several recent efforts to incorporate descriptive weather narratives into hazard prediction pipelines \cite{tan2023application, lai2024bert4st}. Beyond textual processing, transformers have also been adapted for sequential numerical data, enabling unified architectures that learn from temporal patterns and contextual signals in parallel \cite{cholakov2021transformers, lai2024bert4st}. 

These developments set the stage for the proposed dual-stream learning frameworks that bridge structured and unstructured weather data in a unified, predictive system.

%% file: sections/method.tex
This section presents the design and rationale of the poposed dual-stream learning for local wind hazard prediction.

\vspace{-0.5em}
\subsection{Architecture Overview}

\begin{figure}[h]
    \centering
    \vspace{-1.25em}
    \includegraphics[width=0.99\linewidth]{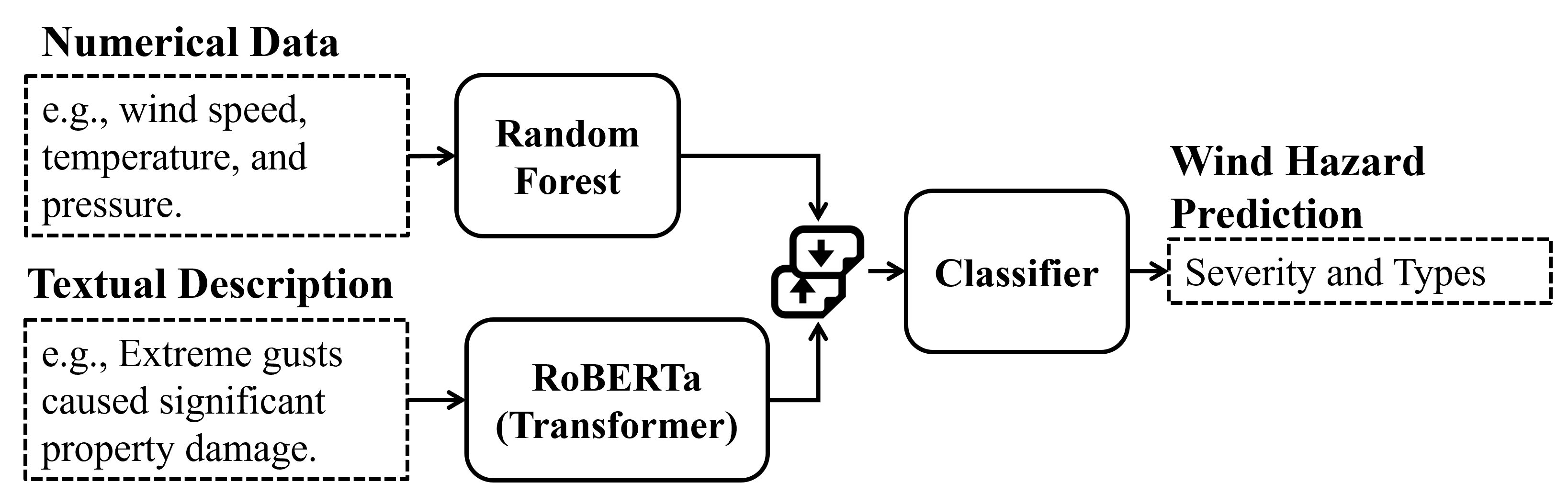}
    \vspace{-1.5em}
    \caption{Overview of the proposed dual-stream architecture. Numerical features are processed by a Random Forest, textual narratives by a RoBERTa encoder, and outputs are fused for localized wind hazard prediction.}
    \vspace{-0.75em}
    \label{fig:architecture}
\end{figure}

The overall model architecture is illustrated in Fig.~\ref{fig:architecture}. The system processes two complementary input modalities: (1) Numerical data, comprising structured meteorological variables such as wind speed, gust intensity, temperature, humidity, and wind direction; and (2) Textual data, consisting of unstructured event narratives describing observed weather conditions.

We propose a dual-stream processing strategy. The numerical features $\mathbf{x}_{\text{num}}$ are processed by a Random Forest (RF) classifier $f_{\text{RF}}(\cdot)$ to model structured atmospheric signals, while the textual narratives $\mathbf{x}_{\text{text}}$ are encoded into semantic feature embeddings by a pretrained RoBERTa transformer~\cite{liu2019roberta} $f_{\text{RoBERTa}}(\cdot)$. 
The respective model outputs are defined as:
\begin{align}
    \mathbf{z}_{\text{RF}} &= f_{\text{RF}}(\mathbf{x}_{\text{num}}), \\
    \mathbf{z}_{\text{RoBERTa}} &= f_{\text{RoBERTa}}(\mathbf{x}_{\text{text}}),
\end{align}
where $\mathbf{z}_{\text{RF}}$ represents the class probabilities from the Random Forest, and $\mathbf{z}_{\text{RoBERTa}}$ denotes the class logits from RoBERTa.

To integrate the complementary strengths of both modalities, a late fusion strategy is employed. The two outputs are concatenated into a unified feature vector:
\begin{align}
    \mathbf{z}_{\text{fused}} = [\mathbf{z}_{\text{RF}}; \mathbf{z}_{\text{RoBERTa}}],
\end{align}
where $[\cdot;\cdot]$ denotes vector concatenation.

The fused vector $\mathbf{z}_{\text{fused}}$ is then passed through a lightweight feedforward neural network $g(\cdot)$, acting as a meta-classifier, to produce the final prediction:
\begin{align}
    \hat{y} = g(\mathbf{z}_{\text{fused}}),
\end{align}
where $\hat{y}$ is the predicted hazard risk level (low or high risk).

During training, the meta-classifier parameters are optimized by minimizing a cross-entropy loss:
\begin{align}
    \mathcal{L} = -\left( y \log \hat{y} + (1 - y) \log (1 - \hat{y}) \right),
\end{align}
where $y$ denotes the ground truth label.

Importantly, both $f_{\text{RF}}(\cdot)$ and $f_{\text{RoBERTa}}(\cdot)$ are pretrained independently and kept fixed during the fusion training stage, ensuring modularity and stability across heterogeneous data sources. Only the meta-classifier $g(\cdot)$ is trained end-to-end during fusion learning.

\subsection{Design Motivation and Advantages}
 
The dual-stream design addresses critical challenges in localized wind hazard prediction. Numerical meteorological data provide precise physical measurements of atmospheric conditions; however, they may miss detailed indicators of risk, such as minor damage reports or unusual local phenomena. Conversely, textual event narratives often encode rich situational context but lack standardized structure.

By jointly leveraging these two data sources, the proposed model achieves several advantages:
% Complementary Feature Capture
(1) Structured and unstructured signals are jointly utilized for complementary information, enhancing predictive richness and robustness.
% Context-Aware Forecasting
(2) Semantic information are extracted from text complements numerical trends, improving early detection of localized hazards.
% Modular Design:
(3) The separate training of Random Forest and RoBERTa modules allows independent optimization and easier system updates as new data becomes available.
% Robust to Missing Data
(4) If one modality suffers from data sparsity or corruption (e.g., missing sensor readings), the complementary modality can still contribute to prediction, increasing system reliability.

Furthermore, the late fusion strategy avoids overfitting risks, by preserving the strengths of each modality until the final decision layer. This structure is particularly well-suited to real-world emergency management applications, where heterogeneous data quality and timeliness are persistent challenges.

%% file: sections/experiments.tex
This section presents the experimental design, including dataset preparation, training procedures, performance evaluation, comparative analysis, and feature importance studies. The objective is to comprehensively assess the proposed dual-stream wind hazard prediction model under realistic operational conditions.

\subsection{Dataset and Preprocessing}

The experimental dataset was collected from the Sioux Gateway Airport weather station~\cite{KSUX}, located in the Midwest United States. It integrates structured meteorological measurements with descriptive event narratives to facilitate multimodal learning. This station provides reliable, high-frequency observations representative of severe wind conditions in rural and tribal regions, making it well-suited for developing and evaluating localized hazard prediction systems.

A total of 10,000 annotated samples were used, with 8,000 randomly allocated for training and 2,000 samples for testing. 
Each sample contains six numerical weather attributes: temperature (tmpf) in degrees Fahrenheit, dew point (dwpf) in degrees Fahrenheit, relative humidity (relh) in percentage, wind direction (drct) in degrees, wind speed (sknt) in knots, and wind gust (gust) in knots. These six features were selected from the original data reports based on expert guidance to capture the most relevant wind-related indicators for risk assessment in operational settings. 
In addition, each sample includes an unstructured Event Narrative field, which summarizes the impact and characteristics of each weather event. An example of an event narrative is ``Extreme gusts caused significant property damage.''

Each of the 10,000 samples is labeled with a binary risk level indicating event severity. 
Low-risk labels correspond to events with minimal or no significant impact, while high-risk labels indicate events associated with notable hazards or property damage. 
Among the 8,000 training samples, 5,850 were labeled as low-risk and 2,150 as high-risk.
Among the 2,000 test samples, 1,467 were labeled as low-risk and 533 as high-risk. 
Prior to training, numerical features were standardized to have zero mean and unit variance. Textual narratives were tokenized using the RoBERTa tokenizer with a maximum sequence length of 128 tokens.

\subsection{Training and Performance}

As discussed, for the textual modality, a pretrained RoBERTa-base model was employed. The maximum token sequence length was set to 128. Training was conducted for 150 epochs using the AdamW optimizer with a learning rate of $3 \times 10^{-5}$ and a weight decay coefficient of 0.01. Mixed precision training was enabled to accelerate convergence. The loss function used for optimization was standard cross-entropy loss.

For the numerical modality, a Random Forest classifier was trained using 100 decision trees, with Gini impurity as the splitting criterion. Each tree was restricted to a maximum depth of 12 to reduce overfitting, and at each node, a random subset of features—equal to the square root of the total number of input features—was considered to promote diversity among the trees. To mitigate potential class imbalance, class priors were automatically adjusted in proportion to the inverse frequency of each class, ensuring fair treatment of both low- and high-risk events during training. The model operated on Term Frequency and Inverse Document Frequency representations derived from the textual component of the dataset. The vocabulary was limited to the top 1,000 most informative terms, including both unigrams and bigrams, while terms appearing in fewer than five documents were excluded to reduce noise and improve generalization. The Gini impurity, defined as \( G = 1 - (p_0^2 + p_1^2) \), where \( p_0 \) and \( p_1 \) represent the proportions of low-risk and high-risk samples within a node, was minimized to produce optimal splits that maximized class separation.

% For the numerical modality, a Random Forest classifier was trained with 100 trees, using Gini impurity as the splitting criterion and a random seed of 42 to ensure reproducibility. The model was built using TF-IDF vectorized text features with a vocabulary limited to the top 1000 terms, including both unigrams and bigrams, and terms appearing in fewer than 5 documents were excluded. Each decision tree in the forest was restricted to a maximum depth of 12, and at each node, a random subset of features (square root of the total number of features) was considered to increase diversity among trees. The \texttt{class\_weight='balanced'} parameter was used to automatically adjust for any class imbalance, helping the model treat both risk levels fairly. The Gini impurity criterion, defined as \(1 - (p_0^2 + p_1^2)\), where \(p_0\) and \(p_1\) are the class proportions in a node, was minimized during training to create splits that led to purer nodes.

To evaluate generalization performance, stratified 5-fold cross-validation was used. In each fold, the model was trained on 80\% of the data and tested on the remaining 20\%. 
Performance was evaluated using standard metrics, including Precision, Recall, F1-Score, Accuracy, and the Area Under the Receiver Operating Characteristic Curve (ROC-AUC). Precision measures the proportion of correctly predicted positive observations to the total predicted positive observations, while Recall measures the proportion of correctly predicted positive observations to all actual positive observations. The F1-Score provides the harmonic mean of Precision and Recall, offering a balanced measure between false positives and false negatives. Overall Accuracy quantifies the proportion of correct predictions out of all predictions made, and ROC-AUC captures the model's ability to discriminate between classes across different threshold settings.

Table~\ref{tab:rf_performance} presents class-level Precision, Recall, and F1-Score results. The proposed model achieves high predictive performance for both risk classes, with particularly strong results for high-risk classification. On the test set of 2,000 samples, all 1,467 low-risk events were correctly identified (Recall = 1.00), and 493 out of 533 high-risk events were correctly classified (Recall = 0.92). Precision for high-risk events reaches 1.00, indicating no false positives. The macro-averaged F1-Score is 0.975, and the overall accuracy of the model is 98\%.

\begin{table}[htbp]
\vspace{-1.5em}
\caption{Performance on Testing Set for Wind Hazard Prediction}
\label{tab:rf_performance}
\vspace{-0.5em}
\centering
\begin{tabular}{lcccc}
\hline
Class & Precision & Recall & F1-Score & Support \\
\hline
Low Risk & 0.97 & 1.00 & 0.99 & 1467 \\
High Risk & 1.00 & 0.92 & 0.96 & 533 \\
\hline
Accuracy & \multicolumn{4}{c}{98\% (on test set of 2,000 samples)} \\
\hline
\vspace{-1.5em}
\end{tabular}
\end{table}

Table~\ref{tab:roc_auc} complements this view by presenting detailed metrics specific to the high-risk class, including a ROC-AUC score of 0.9876, which reflects the model's strong ability to distinguish between high-risk and low-risk cases across varying decision thresholds. This table emphasizes the model’s performance on the more critical high-risk class, which is often the most important in emergency response settings.

\begin{table}[htbp]
% \vspace{-1.5em}
\caption{Additional Evaluation for High-Risk Prediction}
\vspace{-1.0em}
\label{tab:roc_auc}
\centering
\begin{tabular}{lc}
\hline
Metric & Score \\
\hline
Precision (Risk)     & 1.0000 \\
Recall (Risk)        & 0.9250 \\
F1 Score (Risk)      & 0.9610 \\
ROC-AUC Score        & 0.9876 \\
\hline
\vspace{-2.5em}
\end{tabular}
\end{table}

\begin{figure}[htbp]
\centering
\vspace{-1.0em}
\includegraphics[width=0.65\linewidth]{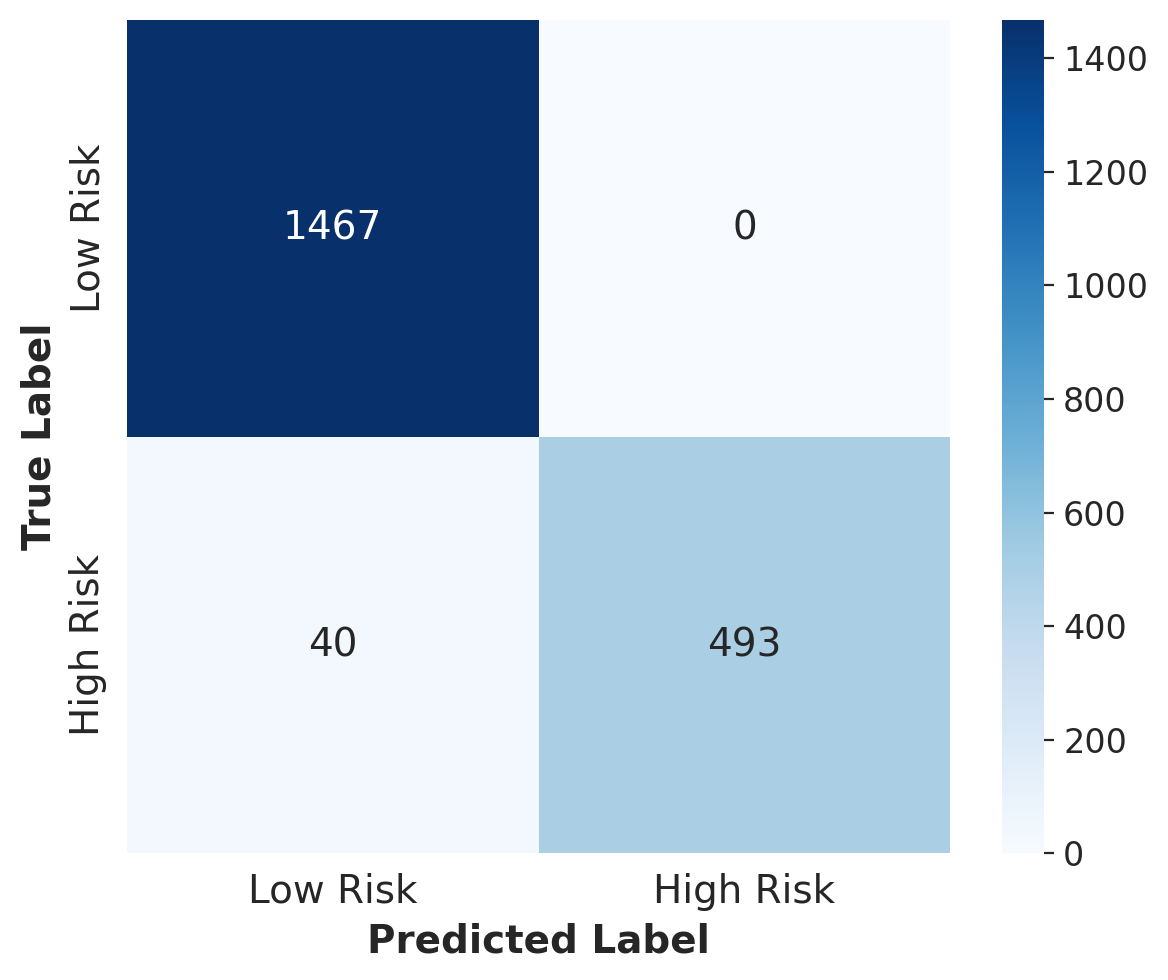}
\vspace{-1.0em}
\caption{Confusion matrix showing model prediction results for low-risk and high-risk events.}
\label{fig:confusion_matrix}
\vspace{-1.0em}
\end{figure}

The confusion matrix in Fig.~\ref{fig:confusion_matrix} further illustrates the classification results. Of the 533 high-risk samples, 493 were correctly classified, while 40 were misclassified as low risk. No low-risk samples were misclassified, reinforcing the model's ability to maintain a low false positive rate, which is essential in real-world emergency response where false alarms can strain limited resources.

\begin{figure}[htbp]
\centering
\vspace{-1.0em}
\includegraphics[width=0.99\linewidth]{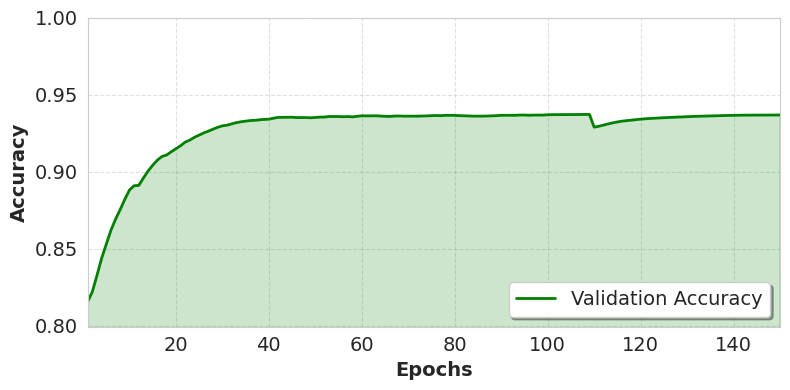} 
\vspace{-2.00em}
\caption{Training and validation accuracy curves over epochs.}
\label{fig:accuracy_curve}
\vspace{-1.0em}
\end{figure}

\begin{figure}[htbp]
\centering
% \vspace{-0.5em}
\includegraphics[width=0.99\linewidth]{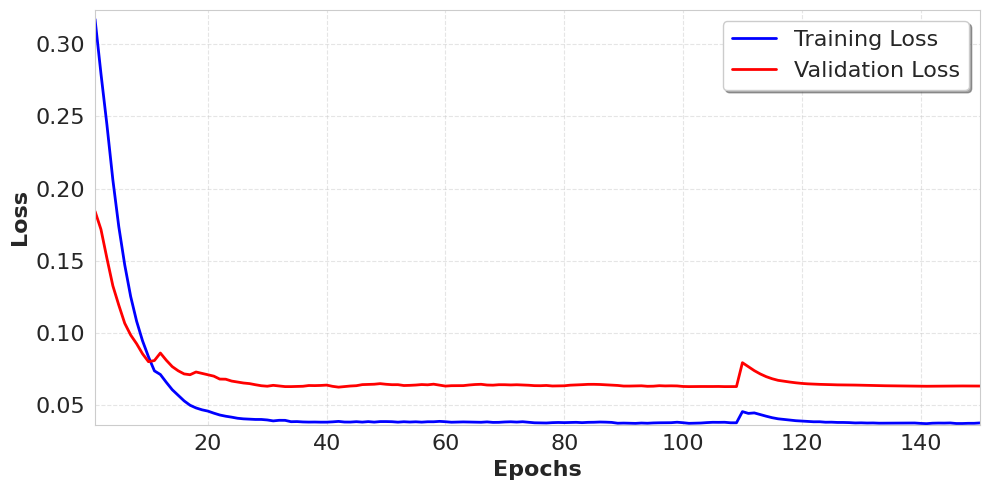}
\vspace{-2.00em}
\caption{Training and validation loss curves over epochs, indicating stable convergence.}
\label{fig:loss_curve}
\vspace{-1.5em}
\end{figure}

The training and validation dynamics over time are visualized in Fig.~\ref{fig:accuracy_curve} and Fig.~\ref{fig:loss_curve}. The accuracy curves show rapid convergence within the first 20 epochs and stable performance across 150 epochs. The loss curves indicate that both training and validation loss decreased sharply at the beginning and remained low, with no sign of overfitting. The slight variation in validation loss near epoch 110 may reflect a minor distributional fluctuation but does not significantly affect performance.

Overall, the results demonstrate that the proposed dual-stream model not only achieves high classification accuracy but also maintains stability and generalization across folds. Its strong performance on high-risk cases, as highlighted in both Tables I–II and the confusion matrix, reinforces its potential for supporting real-time risk assessment in emergency settings.

\subsection{Comparison}

To assess the benefits of multimodal integration, we conducted a comparative and model ablation evaluation against several baseline models. The baselines included traditional machine learning classifiers trained solely on numerical features, as well as a transformer-based model trained exclusively on textual narratives. Specifically, Logistic Regression and Decision Tree classifiers were trained using the structured meteorological variables. A standalone Random Forest classifier was also evaluated to establish a strong numerical-only benchmark. Separately, a RoBERTa-based transformer model was fine-tuned on the event narratives without access to numerical features. Finally, our proposed model, combining Random Forest and RoBERTa outputs through late fusion, was evaluated to highlight the advantages of multimodal learning.

The evaluation results are summarized in Table~\ref{tab:baseline_comparison}. Among the traditional numerical-only models, the Decision Tree achieved an accuracy of 80\% and a macro-averaged F1-Score of 0.76, while the Logistic Regression model attained an accuracy of 73\% and an F1-Score of 0.63. The Random Forest classifier performed substantially better, achieving 98\% accuracy and a 0.97 F1-Score. The transformer-based model trained solely on text achieved an accuracy of 93\% and a 0.90 F1-Score, indicating that textual narratives alone provide meaningful hazard indicators. 
However, the proposed dual-stream model achieved the highest performance, with an accuracy of 98.6\% and a macro-averaged F1-Score of 0.99. These results clearly demonstrate that combining structured and unstructured data sources yields superior predictive capability compared to using either modality alone. The improvement highlights the complementary nature of meteorological measurements and situational narratives in accurately assessing wind hazard risks.

\begin{table}[htbp]
\vspace{-1.5em}
\caption{Model Comparison and Multimodal Ablation Study}
\vspace{-0.5em}
\label{tab:baseline_comparison}
\centering
\begin{tabular}{lcc}
\hline
Model & Accuracy & F1 (Macro) \\
\hline
Logistic Regression (Numerical only) & 73.0\% & 0.63 \\
Decision Tree (Numerical only) & 80.0\% & 0.76 \\
Random Forest (Numerical only) & 98.0\% & 0.97 \\
RoBERTa (Text only) & 93.0\% & 0.90 \\
RF + RoBERTa (Proposed Dual-Stream) & 98.6\% & 0.99 \\
\hline
\vspace{-2.5em}
\end{tabular}
\end{table}

\subsection{Attribute Importance Analysis}

To better understand the decision-making process of the model, feature importance analyses were conducted using two complementary techniques: gradient-based sensitivity analysis and feature ablation studies.

The gradient-based sensitivity analysis measures how small perturbations in the numerical input features affect the predicted risk score. Formally, for a given input $\mathbf{x}$ and model output $\hat{y}$, the gradient $\nabla_{\mathbf{x}} \hat{y}$ indicates the direction and magnitude by which infinitesimal changes in each feature would alter the model's prediction. Features associated with larger absolute gradient values are deemed more influential, as slight variations in their values lead to noticeable changes in the predicted risk level. In practice, gradients were computed for correctly classified high-risk samples to identify which meteorological attributes most significantly impacted hazard assessments. 

Tables~\ref{tab:gradient_analysis} summarize the results of the gradient-based sensitivity. The analysis revealed that wind speed exhibited the highest gradient magnitude, indicating that small fluctuations in wind speed had a strong effect on increasing or decreasing the model’s confidence in high-risk classification. Relative humidity and wind direction also showed non-negligible sensitivities, whereas temperature and wind gust features had comparatively minor impacts.

\begin{table}[htbp]
\centering
\vspace{-1.5em}
\caption{Gradient-based Sensitivity Analysis of Meteorological Features}
\resizebox{0.48\textwidth}{!}{%
\begin{tabular}{|l|c|p{4.2cm}|}
\hline
\textbf{Feature} & \textbf{Gradient Value} & \textbf{Interpretation} \\
\hline
\texttt{tmpf} & $-3.029 \times 10^{-31}$ & Temperature increase slightly reduces high-risk score \\
\hline
\texttt{relh} & $2.386 \times 10^{-31}$ & Humidity increase raises high-risk confidence \\
\hline
\texttt{drct} & $-4.157 \times 10^{-31}$ & Wind direction has small negative influence \\
\hline
\texttt{sknt} & $-7.186 \times 10^{-31}$ & Wind speed has the strongest gradient influence on risk score \\
\hline
\texttt{gust} & $2.826 \times 10^{-32}$ & Wind gust has minimal positive effect on prediction \\
\hline
\end{tabular}%
}
\vspace{-1.0em}
\label{tab:gradient_analysis}
\end{table}

In parallel, an attribute ablation study was performed by systematically removing each numerical feature from the input and evaluating the change in the classification confidence. For each feature, the model's output was recalculated after zeroing out that feature while keeping all others unchanged. If the removal of a feature led to a significant drop in classification confidence, the feature was considered critical. 

Table~\ref{tab:ablation_analysis} summarize the results of the attribute ablation study. The ablation results showed that removing wind direction information caused the most substantial decrease in high-risk classification confidence, suggesting that wind direction plays a pivotal role in distinguishing between events. 

Interestingly, although wind speed was the most sensitive feature in the gradient study, its removal during ablation had a smaller impact, indicating that while small changes in wind speed strongly influence model predictions, the model can compensate when the entire feature is missing. Conversely, wind direction showed a lower gradient magnitude—suggesting it exerts a more stable, background-level influence—but its removal caused a substantial drop in classification confidence. This contrast reveals how the two analyses provide complementary insights: gradient analysis captures local sensitivity to small perturbations, while ablation highlights global feature necessity for decision robustness.

\begin{table}[htbp]
% \vspace{-1.0em}
\centering
\caption{Feature Ablation Analysis and Impact on High-Risk Prediction}
\resizebox{0.48\textwidth}{!}{%
\begin{tabular}{|l|c|p{4.2cm}|}
\hline
\textbf{Feature} & \textbf{Ablation Impact} & \textbf{Interpretation} \\
\hline
\texttt{tmpf} & 0.00 & Temperature: No dominating effect in classification confidence when removed \\
\hline
\texttt{relh} & 0.00 & Humidity: No dominating effect in classification confidence when removed \\
\hline
\texttt{drct} & 0.76 & Wind direction: Critical and large drop in classification confidence when removed \\
\hline
\texttt{sknt} & 0.00 & Wind speed: No dominating effect in classification confidence when removed despite strong gradient \\
\hline
\texttt{gust} & 0.00 & Wind gust: No dominating effect in classification confidence when removed \\
\hline
\end{tabular}%
}
\vspace{-1.5em}
\label{tab:ablation_analysis}
\end{table}

\subsection{Interpretability in Wind Hazard Prediction}

% \xz{Dr. Huang, this is application focused discussion, I believe they can be improved to be more domain accurate}
Attribute importance analyses play a critical role in enhancing the interpretability and practical applicability of the proposed wind hazard prediction model. In high-stakes domains such as emergency management, it is not sufficient for a predictive model to only achieve high accuracy; it must also provide transparent and explainable reasoning to support informed decision-making by human operators. 

The gradient-based sensitivity attribute analysis and ablation study jointly offer insights into the internal behavior of the model, revealing how different meteorological elements contribute to risk assessment. By identifying wind speed and wind direction as key drivers of high-risk predictions in the study region, the model’s outputs become more understandable and verifiable to domain experts, who can cross-reference these findings with established meteorological knowledge and local wind damage records. 
Improved interpretability also builds trust among emergency management personnel and other stakeholders, including community residents. When stakeholders can understand the basis of a model’s decisions, they are more likely to adopt and rely on its outputs during critical operations such as issuing warnings or allocating emergency resources. Moreover, interpretability facilitates error analysis and model refinement by making it easier to diagnose potential failures or unexpected behaviors in the system. 

Additionally, feature importance findings can guide the prioritization of sensor maintenance and manual validation efforts. For example, knowing that wind direction has a disproportionately high impact on prediction outcomes suggests that maintaining the accuracy of wind direction sensors should be a priority. Similarly, manual verification efforts could be focused on events where key features show abnormal values, enabling more efficient quality control and system robustness.

%% file: sections/conclusion.tex
This work presents an interpretable dual-stream learning framework for wind hazard prediction, targeting the unique needs of vulnerable and underserved communities. By integrating structured numerical weather features with unstructured event narratives, the proposed system achieves robust, localized risk assessment. Experimental results demonstrate that the fused model significantly outperforms traditional single-modality baselines across multiple metrics. In addition, feature sensitivity and ablation analyses provide valuable interpretability, supporting informed decision-making for emergency management. The framework is designed with operational deployment in mind and can be integrated into real-time decision-support tools. Future work will explore extension to additional modalities such as satellite imagery and incorporate real-time integration with GIS-based emergency coordination systems.